\documentclass[conference]{IEEEtran}
\usepackage[latin1]{inputenc}
\usepackage{makeidx}          
\usepackage{graphics,graphicx}          
\usepackage{amssymb,amsmath,amsthm}  
\usepackage{subfigure}
\usepackage{textcomp}
\usepackage[margin=19mm]{geometry}
\usepackage{algorithm}
\usepackage{algorithmic}
\usepackage{float}
\usepackage{color}
\usepackage{verbatim}
\usepackage{enumerate}

\IEEEoverridecommandlockouts

\title{Towards Deep Machine Reasoning: a Prototype- based Deep Neural Network with Decision Tree Inference
\thanks{Plamen Angelov, and Eduardo Soares are with the School of Computing and Communications, Lancaster University, Lancaster, LA1 4WA, UK.  E-mails: p.angelov@lancaster.ac.uk; e.almeidasoares@lancaster.ac.uk.}}

\author{Plamen Angelov, Eduardo Soares}

\begin{document}

\maketitle

\begin{abstract}

In this paper we introduce the DMR -- a prototype-based method and network architecture for deep learning which is using a decision tree (DT)- based inference and synthetic data to balance the classes. It builds upon the recently introduced xDNN method \cite{angelov2019explainable} addressing more complex multi-class problems, specifically when classes are highly imbalanced. DMR moves away from a direct decision based on all classes towards a layered DT of pair-wise class comparisons. In addition, it forces the prototypes to be balanced between classes regardless of possible class imbalances of the training data. It has two novel mechanisms, namely i) using a DT to determine the winning class label, and ii) balancing the classes by synthesizing data around the prototypes determined from the available training data. As a result, we improved significantly the performance of the resulting fully explainable DNN as evidenced by the best reported result on the well know benchmark problem Caltech-101 surpassing our own recently published "world record". Furthermore, we also achieved another "world record" for another very hard benchmark problem, namely Caltech-256 as well as surpassed the results of other approaches on Faces-1999 problem. In summary, we propose a new approach specifically advantageous for imbalanced multi-class problems that achieved two world records on well known hard benchmark problems and the best result on another problem in terms of accuracy. Moreover, DMR offers full explainability, does not require GPUs and can continue to learn from new data by adding new prototypes preserving the previous ones but not requiring full retraining.         
\end{abstract}

\section{Introduction}

In this paper we introduce a new deep neural network (DNN) which departs from the amorphous and highly abstract, \textit{"black box"} model structure towards deep machine reasoning (DMR) architecture. This is based on the following principle differences from the traditional approach: i) use of prototypes as the core of the method; ii) use of a DT for decision making (class labeling) instead of a flat "\textit{winner takes all}" type function; iii) using similarity as a measure of association to prototypes; iv) possibility to express the method in a form of human-interpretable IF-THEN rules with partial degree of satisfaction and to visualise by Voronoi tessellation or by prototypes.

The staggering increase of the amount and complexity of the data sets and streams led to a move from rule-based systems (fuzzy, Bayesian inference, Markov decision processes, using Q tables in reinforcement learning, case base reasoning, etc.) towards DNN which have proven their efficiency in a number of problems ranging from speech, image recognition and language translation to games \cite{goodfellow2016deep}. This abundance of data led, however, to the temptation to shortcut from data to the solutions driven entirely by the accuracy and ignoring the depth of understanding the problem at hand, and getting insights.

In DRM we make use of the strong properties of the DNN and add new mechanisms to address their shortcomings. For example, the DNN are very efficient feature extractors, especially for image processing problems \cite{goodfellow2016deep}. We use this in DRM  and we also use layered structure/architecture. We further benefit from the transfer learning approach \cite{pratt1993discriminability}. 

Traditional classifiers assume balanced classes, but in practice classes are usually (highly) imbalanced. For example, in fault detection and identification the amount of data about the faulty cases are usually significantly smaller than the amount of data for "normal" operation \cite{costa2015fully}. In social applications, for example, this leads to possible un-fairness \cite{soares2019fair} when the data is highly imbalanced with dominating class(es) and minority class(es). 

Finally, traditional statistical modelling is heavily influenced by averages and starts with assumptions about the data distributions which are then put to a test by parametrisation \cite{hastie2009elements}. We take the opposite approach starting with the observed data samples and generalise from these local densities and global multivariate generative distributions. These empirically derived distributions have discrete and continuous form \cite{angelov2019empirical}. Their discrete forms are exact while the continuous forms which are needed for the inference are local estimates.

Prototype-based models have demonstrated their high efficiency, e.g. the discriminative models such as kNN \cite{zhang2017learning}, SVM \cite{suykens1999least}, less so RBF \cite{er2002face} and LVQ \cite{kohonen1996lvq}. The latter two are also good in terms of explainability \cite{angelov2019explainable}. Explianability is undoubtedly, the Achilles heel of the DNN and the solution we propose is to have a synergy between reasoning and learning rather than the current dichotomy.

In this paper we offer a new deep learning architecture and method that builds upon our recently introduced xDNN \cite{angelov2019explainable} method by adding two important novelties, namely: i) using a DT to determine the winning class label, and ii) balancing the classes by synthesising data around the prototypes determined from the available training data. 

We validated the new DMR method on three well known benchmark problems, namely Faces-1999, Caltech-101 and Caltech-256. Both Caltech problems are very hard and there is a public record of the best results achieved so far \cite{he2015spatial}. We surpassed one of them (Caltech-101) with xDNN already \cite{angelov2019explainable}. With DMR we surpass our own xDNN "world record". Furthermore, we also surpassed the best record on Caltech-256 as well as on Faces-1999 problems. Moreover, DMR does not require GPUs, computationally lean and can continue to train for new data without the need for full re-training. 

The remainder of the paper is organised as follows: Section II introduces the concept and novelties of the proposed approach. Section III presents the proposed architecture used during the training phase. Section IV outlines the learning procedure, section V introduces the architecture of the DMR used during the validation phase. Section VI illustrates explainability of DMR in terms of IF-THEN rules. Numerical experiments are presented in the Section VII, results are analysed in Section VIII and the paper is concluded with Section IX.       

\section{Concept and novelties of the proposed approach}

The problem we consider in this paper is to design a classifier with deep architecture that is \textit{explainable-by-design} due to the use of prototypes \cite{angelov2019explainable}. Prototypes are a small subset of the training data that are highly representative. This is because they are the local peaks of the distribution \cite{angelov2019empirical}. 

Let us denote the training data set of points by $x = \left \{ x_1,..., x_{N} \right \} \in \mathbb{R}^n$ with corresponding class labels $y_1,..., y_{C}$  $\in \left \{ 1,..., C \right \} $. Here, $N$ is the number of training data samples and $n$ is their dimensionality (number of features); $C$ is the number of  classes. DMR starts by selecting a set of descriptive prototypes $\pi$ $\in X$ $\subset P$ for each class/per class, $M_j$ is the total number of prototypes of class $j$; $M_j =|P_j|$;
$M=\sum_{j=1}^ {C}M_j$. Notice that $M_j>1$ for $ \forall  j$, i.e. we usually consider more than a single prototype per class .
The prototype extraction process (which can be both, offline and online) is described in more detail in \cite{angelov2019empirical},\cite{angelov2019explainable}. At the heart of practically all prototype-based methods is the concept that the prototypes of class $C$ are designed to be close to many training points of class $C$ and far from training points of the other classes. As pointed out in \cite{angelov2019explainable} "This idea captures the sense in which the word \textit{prototypical} is commonly used". 

The power of prototype-based approaches stems from the fact that they are\textit{ explainable-by-design} \cite{soares2019fair}, easy to understand by the users because they represent samples of the training data, e.g. images. They can be used for classification. Any new data sample with unknown label, $x \in \mathbb{R}^n$ can be associated with the nearest prototype from the sets $P_1,P_2,...,P_C$; $P=P_1 \cup P_2 \cup ... \cup P_C$.

\begin{equation}
L(x) = \operatorname*{argmin}_{ x \in X}\operatorname*{min}_{ \pi \in P}d(x,\pi).
\label{label}
\end{equation}

\subsection{Decision Tree layer}

 In traditional DNN, the decision is flat, \textit{en bloc} in the form of a single stage \textit{"winner takes all"} function as in eq. (\ref{label}) and is the last layer of the network. In xDNN \cite{angelov2019explainable} we also followed this popular decision concept, but split it into two stages: i) per class winner, and ii) across classes global decision. 
 In DMR, similarly to xDNN \cite{angelov2019explainable} the decision mechanism is part of the architecture used for validation of the results because the training is per class and no decision for the class label is needed during the training.
 In this paper, the proposed DMR is using a multi-layer DT formed by pairwise comparison of top two classes in terms of minimum error in training as detailed in Section V and Fig. (\ref{Fig7}). The reason the result is significantly different is that the Voronoi tessellation regions of the data clouds that are formed around each prototype (local zones of influence) are significantly different when binary decisions are made.

\subsection{Balancing classes through synthesising training data strategically}

The second innovation of the proposed method is related to the balancing of the classes. We achieve this by synthetic data augmentation. In this paper we propose a different approach from our recently published one \cite{gu2019self} for synthesising data for highly imbalanced classification problems. The differences are that in this paper we synthesise data around prototypes which makes these synthetic data more likely to have the same class as the prototype. The method starts by identifying a population of pairwise neighbouring data samples from minority classes around prototypes. Then, it imposes a Gaussian disturbance on these data samples, and, finally, it generates synthetic samples by creating linear interpolations between these extrapolations. A further difference from our recent method \cite{gu2019self} is that in this paper we use the standard deviation, $\sigma$ as a radius of influence around the prototype rather than absolute distance of first order. We then augment the training data set with this synthetically generated data set as shown in Fig. (\ref{fig1}), see the augmented prototypes layer.

\section{Architecture of the proposed DMR approach (during the training phase)}

The architecture of the proposed classifier can be represented as a multi-layered DNN with a very clear semantic and functional meaning by design. The architecture for the training and for the validation phases are different as detailed in Figs. \ref{fig1} and \ref{Fig7}. The training phase is performed per class (except the last layer) and includes the following layers:

\begin{figure*}
  \centering
  \includegraphics[width=0.7\linewidth]{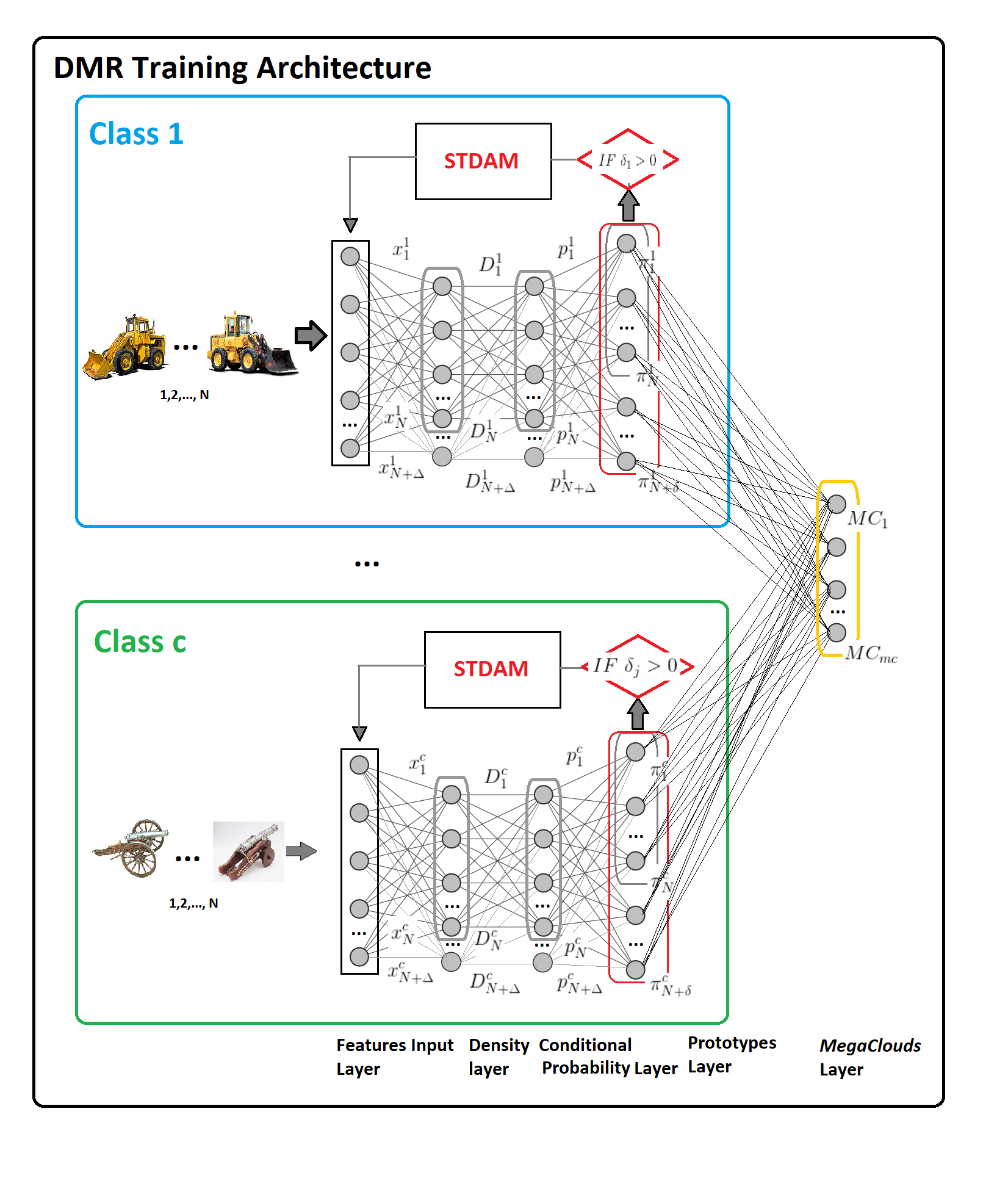}
  \caption{DMR Architecture during the training phase (STDAM stands for Synthetic Training Data Augmentation Mechanism). }
  \label{fig1}
\end{figure*}%

\begin{enumerate}
    \item \textbf{Input (features) layer}
    
This is the first layer which defines the data space. The number of inputs is determined by the nature of the problem that the data describe. In many problems these are clearly known physical or biomedical variables, e.g. velocities, pressure, temperature, etc. In image processing problems traditionally size, shape of objects or HoG \cite{mizuno2012architectural} were used as well as more abstract methods like GIST \cite{solmaz2013classifying}. More recently, convolutional neural networks (CNN) like AlexNet \cite{krizhevsky2012imagenet}, VGG--VD--16 \cite{simonyan2014very}, Inception \cite{szegedy2015going}, ResNet \cite{he2016deep}, Inception--Resnet \cite{szegedy2017inception} have proven to be very efficient to encode images and represent them as a highly abstract vector of the outputs from the Fully Connected Layer (FCL). The proposed DMR architecture is agnostic to the source of the features vector that the input layer represents. It can be any of the above. In this paper without any loss of generality we use a $1\times4096$ dimensional vector formed by the outputs from the first FCL from a VGG--VD--16 pre-trained on Imagenet \cite{deng2009imagenet}. 

\item \textbf{Data density layer}

This layer is composed of neurons who's activation function represent the data density, \textit{D} defined by a Cauchy function \cite{angelov2019empirical}:

\begin{eqnarray}
D(x) = \frac{1}{1+\frac{||x-\mu||^2}{\sigma^2}}, 
\label{eqS}
\end{eqnarray}

where $D$ is the density, $\mu$ is the global mean, and $\sigma$ is the variance. 
In \cite{angelov2019empirical} it was demonstrated theoretically that starting from the mutual proximity of the data samples in the data space and using Euclidean (or Mahalanobis) type distance \textit{D} takes the form of a Cauchy function. Moreover, data density can be updated recursively as detailed in \cite{angelov2012autonomous}. The value of the data density, $D$ represent the closeness to the mean and is in the range $0<D\leq1$. It obtains its maximum (of 1) when $x=\mu$. $D$ is indicative for the centrality of a data sample and its suitability to be a prototype due to its proximity to other data samples. 

\item \textbf{Conditional probability layer}

The conditional probability can be estimated from the empirically observed data as described in \cite{angelov2019empirical} where it is also called $typicality$ $\tau$. It can be given by eq. (\ref{tau}). The integral of $\int_{-\infty }^{\infty}p(C|x) dx =1$ same as for the pdf \cite{angelov2019empirical}, but it is multi-modal: 

\begin{equation}
p(C|x)= \frac{\sum_{i=1}^C N_i D(x^c)}{\sum_{i=1}^C N_i \int_{-\infty}^\infty D(x^c)dx}
\label{tau}
\end{equation}

\noindent where $N_i$ denotes the number of data samples associated with (support of) the $i-th$  \textit{data cloud}, $\sum_{i=1}^C$; $N_i = N$.
Notice that since p(C$|$x) is empirically derived \cite{angelov2019empirical} it is not constrained by any \textit{prior} assumptions about the data distribution type or even about the random or deterministic nature of the data. This is clearly more realistic in comparison with the common approach which (for theoretical convenience) assumes randomness and independence of the features of the experimentally observed data which is usually far from the reality. 

\item \textbf{Prototypes layer}

The next layer consists of prototypes, $\pi$. This is the core layer of the proposed DMR architecture. This layer is responsible to provide \textit{explainable-by-design} model.  Prototypes are the local peaks of the data density (and, respectively, local peaks of the conditional probability, eq. (\ref{tau})) identified in the previous layers/stages. The proposed DMR algorithm absorbs the new data samples by assigning them to the nearest prototype: 

\begin{equation}
j^*=\operatorname*{argmin}_{ i=1,..,N; j=1,..,M} |x_{i}-\pi_{j}| 
\label{nearest}
\end{equation}

In this way, each prototype forms a $cloud$ of data that it represents. These \textit{"data clouds"} form Voronoi tessellation, illustrated in Fig.\ref{Fig2} 

\begin{figure}[h]
	\begin{center}
		{\includegraphics[scale=0.33]{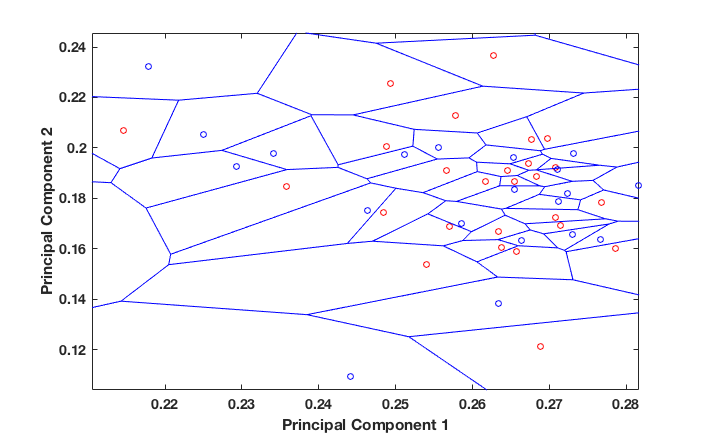}}
		\caption{Identified prototypes -- Voronoi Tessellation. The blue circle represents "class 1" and the red circle denotes "class 2".  } \label{Fig2}
	\end{center}
\end{figure}
\vspace{8pt}

The prototypes are independent from each other. Therefore, one can change the structure by adding a new prototype without influencing the other already existing prototypes. In other words, the proposed DMR network is highly parallelizable and suitable for dynamically evolving applications with non-stationary data streams and evolving data patterns where new prototypes may be added if the data pattern requires this. The proposed DMR network is trained \textit{per class} forming a set of prototypes \textit{per class}. Therefore, all the calculations are done for each class separately. New prototypes are added to this layer when the following condition is met \cite{angelov2019empirical}:

\begin{equation}
\begin{split}
\textit{\text{IF }}(D(x)\geq\max_{j=1,..,M}D(\pi_j))~~\\
\textit{\text{OR }}~~(D(x)\leq \min_{j=1,..,M}D(\pi_j))\\
\textit{\text{ THEN }} (add~a~new~data~cloud ~(j \leftarrow j+1))\label{if}
\end{split}
\end{equation}

If that is the case, then the vector of features of the current training data sample becomes a new prototype, $\pi_{j+1}$ forms a new \textit{data cloud} \cite{angelov2019explainable}. 
\item \textbf{ Synthetic data augmentation}

This mechanism is not a separate layer, but a feedback process that gets information from the prototypes layer, augments the training data set (in the form of synthetically added features vectors close to the existing prototypes) and expands the size of the prototypes layer by balancing the amount of prototypes per class. This mechanism is one of the two novelties of the proposed approach in comparison with our recent xDNN \cite{angelov2019explainable} method. The rationale for and the main functionality of this mechanism has been described in Section II.B. In fact, this is an augmentation of the amounts of training data (by augmenting $N$ to $N+\Delta$ made by feeding back the information from the prototypes layer. As a result, the size of the prototypes layer is expanded (by $\delta$) so that the number of prototypes per class is being balanced. This is visualised in Fig. (\ref{fig1}) where the red solid rectangle includes the black dotted one (original prototypes) but also adds prototypes which result from adding synthetic training data. 

\item \textbf{ \textit{MegaClouds} layer}

This is the final layer of the training architecture. Unlike the previous layers it is cross-class. At this layer prototypes from all classes are put together and once this is done all the adjacent \textit{data clouds} that have the same class label are combined into \textit{mega-clouds}, see Fig.(\ref{Fig3}). Notice that the number of \textit{megaclouds}, $i = 1,2,...,MG$ is significantly smaller than the number of prototypes, ($MG<<M$) and the interpretability improves significantly.

\begin{figure}[h]
	\begin{center}
		{\includegraphics[scale=0.4]{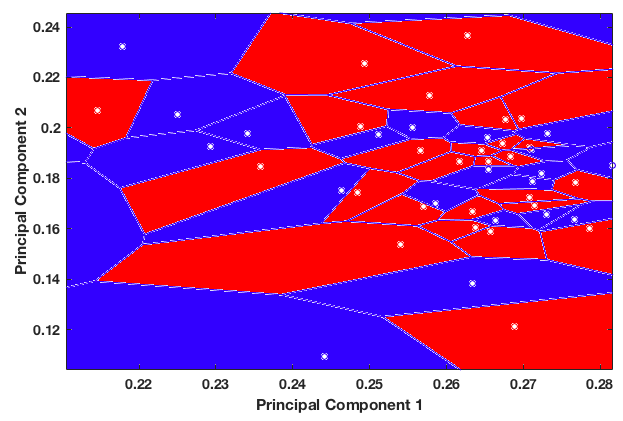}}
		\caption{\textit{Mega-Clouds} are result of merging adjacent \textit{data clouds} which has the same class label.   } \label{Fig3}
	\end{center}
\end{figure}
\vspace{8pt}

\end{enumerate}

\section{Learning Procedure}

The learning of DMR is summarised below by the following pseudo-code. The proposed architecture is feed-forward with the exception of the synthetic data augmentation mechanism which feeds back form the prototype layer back to the input layer. The proposed method can work both, in a batch mode as well as on a per sample basis, online.

\vspace{6pt}
\hrule
\vspace{6pt}
\textbf{DMR: Learning Procedure}
\vspace{3pt}
\hrule
\vspace{4pt}
\begin{algorithmic}[1]

    \STATE Read the first feature vector sample $x_i$ of class $c$;
    \STATE Standardise and normalise the data as detailed in \cite{angelov2019empirical}
	\STATE Set $i \leftarrow 1;  j \leftarrow 1; \pi_1 \leftarrow x_i; \mu  \leftarrow  x_1; N \leftarrow 1$
	\STATE \textbf{FOR} $i$ = 2, ...
	\STATE ~~ Read $x_i$;
	\STATE ~ Calculate $D(x_i)$ and $D(\pi_j)$ $(j=1,2,...,M)$ according to eq. (\ref{eqS});
	\STATE ~~ \textbf{IF} eq. (\ref{if})  holds
	\STATE ~~~~ Create new prototype: $j \leftarrow j+1; \pi_{j} \leftarrow x_{i}; N \leftarrow N+ 1$ 
	\STATE ~~ \textbf{ELSE}
	\STATE ~~~~ Search for the nearest prototype according to eq. (\ref{nearest});
	\STATE ~~~~ Update the nearest prototype as:\\
           ~~~~$N \leftarrow N+1;$ ~~ \\
           ~~~~$\pi_{j} \leftarrow \frac{N_{j}}{N_{j}+1}\pi_{j}+\frac{N_{j}}{N_{j}+1}x_{i};$
	\STATE ~~~~ Balance the number of prototypes through synthetic data augmentation mechanism detailed below;
	\STATE ~~ \textbf{END}
	\STATE \textbf{END}	
\end{algorithmic}
\vspace{4pt}
\hrule
\vspace{6pt}

~~
\vspace{10pt}
\hrule
\vspace{6pt}
\textbf{Synthetic Data Generation}
\vspace{3pt}
\hrule
\vspace{4pt}
\begin{algorithmic}[1]

   	\STATE \textbf{FOR} $j$ = 1,2,...,C \textbf{DO} 
    \STATE ~~Calculate the amount of synthetic data samples needed to balance the pair of classes $j$ and $j+1$: $\delta =M_j-M_{j+1}$. 

	\STATE ~~\textbf{UNTIL} $\delta=0 $ \textbf{DO}
	\STATE ~~~~ $k=1$
	\STATE ~~~~ Randomly select a pair of neighbouring data samples $(p_k,q_k )^*$ from the 0.3$\sigma$ zone around the prototype from the minority class; \\
	\STATE ~~~~ Apply Gaussian disturbance to $(p_k,q_k )^*$  by eq. (\ref{eqGaussian}) and obtain $(\hat{p_k},\hat{q_k})^*$ \cite{freudenberg2010stabilization}; \\
	
		\begin{eqnarray}
(\hat{p}_k,\hat{q}_k)^*= (p_k + g_p, q_k +g_q)^*_k
\label{eqGaussian}
\end{eqnarray}
	
	\noindent where $g_p=[g_{p,1},g_{p,2},...,g_{p,R} ]^T$ and $G_q=[g_{q,1},g_{q,2},...,g_{q,R}]^T$ are two $R$ dimensional randomly generated vectors sampled from the Gaussian distributions, $g_{p,l},g_{q,l}\sim\mathbb{N}(0,\sigma) (l=1,2,...,R)$ with $\sigma$ being the standard deviation.

	\STATE ~~~~ Create random interpolation $\rho_k$ between $(\hat{p}_k,\hat{q}_k)^*$ as follows \cite{gu2019self}: \\
	
	\begin{eqnarray}
\rho_k = \alpha^T_k\hat{p}_k+ (1-\alpha_k)^T\hat{q}_k
\label{eq4}
\end{eqnarray}

\noindent where $\alpha_k = [\alpha_{k,1},\alpha_{k,2},...,\alpha_{k,R}]^T$ is a $R$ dimensional random vector, elements of which follows the uniform distribution within the range [0,1]. 
   \STATE ~~~~$k \leftarrow k+1$
  	\STATE ~~\textbf{END} \textbf{UNTIL}
\STATE \textbf{END} \textbf{FOR}

\end{algorithmic}
\vspace{4pt}
\hrule
\vspace{6pt}

\section{Validation Architecture}

\begin{figure*}[h]
	\begin{center}
		{\includegraphics[scale=0.38]{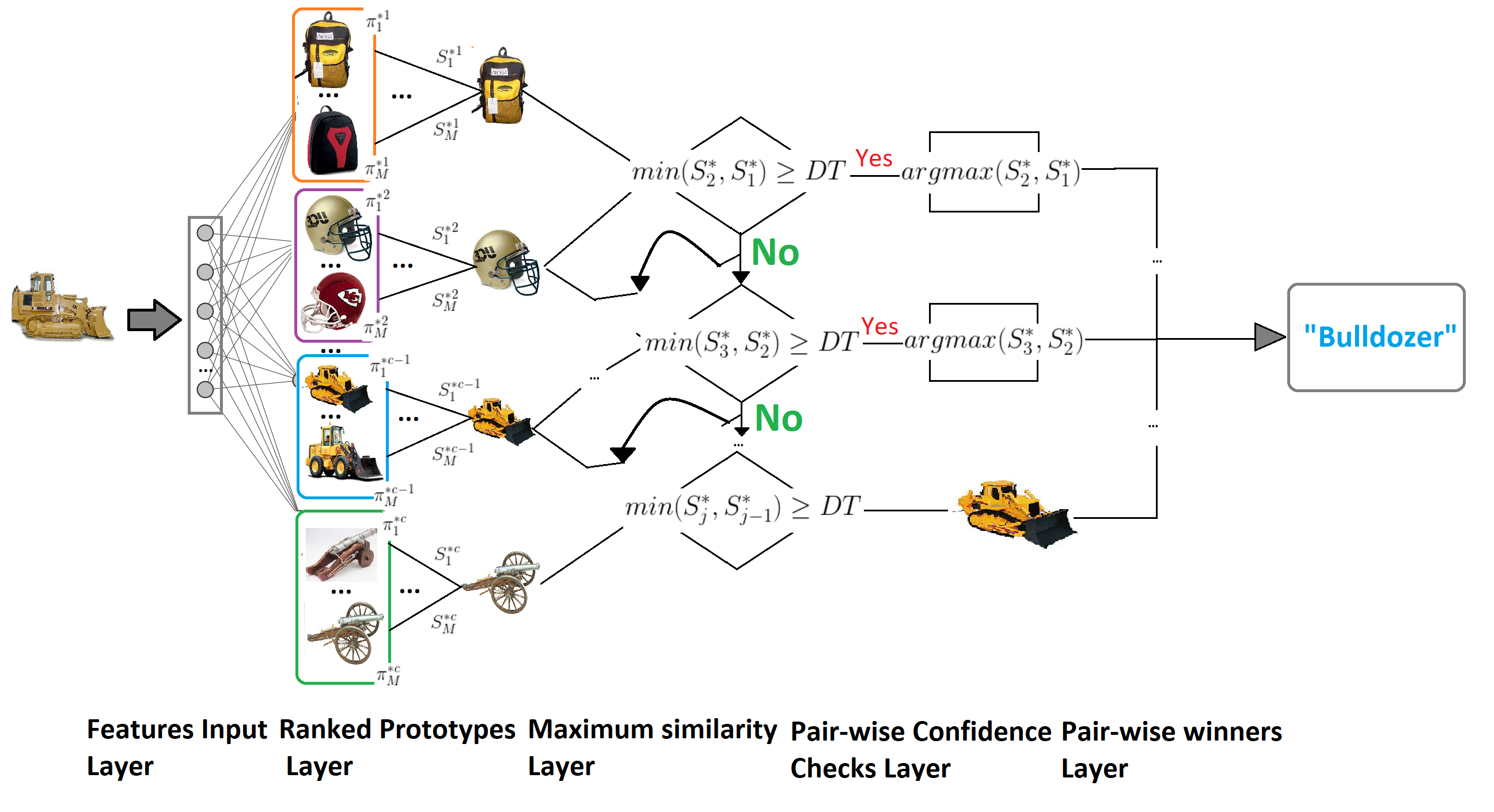}}
		\caption{Architecture for the validation process of the proposed DMR approach.} \label{Fig7}
	\end{center}
\end{figure*}

The architecture of DMR for the validation phase (see Fig. \ref{Fig7}) has the following layers.

\begin{enumerate}

\item{\textbf{Input (features) layer}}
The first layer is exact the same as in the training phase and has been described in section III.

\item \textbf{Ranked prototypes layer}

In this layer we rank order all the prototypes in terms of minimum error during the training. Then we organise them in overlapping pairs: we start with the top two prototypes (providing smaller error) and then the pair of the second best and the third; further on, the pair of the third and the forth, etc. In this way, all prototypes take part twice except the best one and the worst one, see Fig. (\ref{Fig7}). The output of this layer is the degree of similarity, $S$ between the unlabeled data sample and the respective prototype. The activation functions of the neurons of this layer are defined as follows: 

\begin{eqnarray}
S_j=S(x_i,\pi_j) = \frac{1}{1+\frac{(x-\pi_j)}{\sigma^2_j}}, 
\label{eq1}
\end{eqnarray}

where $j=1,2,...,M; i=1,2,...,N$. 
It is easy to see that for similarity we use the same Cauchy function as the data density, eq. (\ref{eqS}).

\item \textbf{Maximum similarity layer}

Each neuron of this layer is performing a simple max operation over the pair of similarity values that are coming form the previous layer, namely:

\begin{eqnarray}
S^*_{j,j-1} = max(S_{j-1},S_j) 
\label{ineq9}
\end{eqnarray}

The winner goes forward.

\item \textbf{Pair-wise confidence checks layer}

In this layer we check if the confidence in the best of the two potential outcomes is high enough. In this paper we use a threshold,  $Thr$=0.9, which means 90\% similarity of the new, unlabeled data sample to any prototype. The neurons of this layer are linked between each other forming a competitive layer. This link is activated if the confidence check fails (see Fig. 2).
The flow of the information to the next layer is conditional on the outcome from the confidence check. First, the top two pairs of prototypes are checked. If the winner surpasses $Thr$ it is the winner. Otherwise, the flow goes down to the next pair (in the same layer of the network, the key Fig. \ref{Fig7} is closed) and so on.

\vspace{6pt}
$IF$  ($min(S^*_{j,j-1},S^*_{j-1,j-2}$)$ \geq Thr$) $THEN$  ($Step$ 4)
\\  $ELSE$  ($Step$ 3 )

\vspace{6pt}

\item \textbf{Pair-wise winners layer}

Pair-wise decisions are made to determine the winning prototype form the candidate pair $(S^*_{j-1},S^*_j)$, which passed the confidence check in the proceeding layer. 

\begin{eqnarray}
Label =argmax (S^*_{j,j-1},S^*_{j-1,j-2})
\label{ineq1}
\end{eqnarray}

\end{enumerate}

\section{Explaining the DMR network as a set of IF...THEN rules}

One of the main advantages of the proposed DMR approach is that it is \textit{explainable-by-design} and can be represented, for example, in the form of IF...THEN rules \cite{angelov2012autonomous}. People can easily understand rules and prototypes. These are often easy to visualise, e.g. in case of images and can also be expressed as a set of linguistic rules as follows:

\vspace{8pt}

\begin{figure}[H]
	\begin{center}
	IF (Image $\sim$ {\includegraphics[scale=0.2]{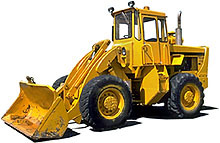}}) THEN "Bulldozer"
		\label{drb_class}
	\end{center}
\end{figure}

\noindent where $ \sim $ denotes "similar to"; it can also be seen as a fuzzy degree of membership. One rule per prototype can be formed. All rules per class can be combined together using logical OR, also known as disjunction or S-norm:

\begin{figure}[H]
	\begin{center}
	IF (Image $\sim$ {\includegraphics[scale=0.2]{Fig5.jpg}})  OR \nonumber (Image $\sim$ {\includegraphics[scale=0.15]{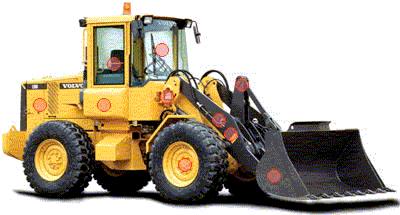}})OR\\ ... OR  (Image $\sim$ {\includegraphics[scale=0.25]{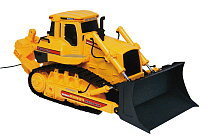}}) \nonumber THEN "Bulldozer"
		\label{drb_class}
	\end{center}
\end{figure}

\section{Numerical Experiments}

We validated our proposed approach, DMR using several complex, well-known image classification benchmark data sets (Faces-1999, Caltech-101, and Caltech-256). Description of the data sets are given below:

\subsubsection{Faces-1999}

The Faces-1999 data set \cite{weber1999caltech} contains 450 frontal real faces images from 27 different people. This data set is highly unbalanced.

\subsubsection{Caltech-101}

The Caltech-101 data set \cite{griffin2007caltech} contains 9144 images in divided into 102 categories(one background). The Caltech-101 dataset is highly unbalanced and is widely used as bench marking data set.

\subsection{Caltech-256}

Caletch-256 has 30,607 images divided into 257 object categories (one of which is the background) \cite{griffin2007caltech}. 

\subsection{Performance Evaluation}

The performance of the classification methods is usually evaluated based on their accuracy index which is defined as follows:

\begin{eqnarray}
ACC(\%) = \frac{TP+TN}{TP+FP+TN+FN},
\label{acc}
\end{eqnarray}

\noindent where $TP, FP, TN, FN$ denote true and false, negative and positive, respectively.

All the experiments were conducted with MATLAB 2018a using a personal computer with a 1.8 GHz Intel Core i5 processor, 8-GB RAM, and MacOS operating system. The classification experiments were executed using 10-fold cross validation under the same ratio of training-to-testing (80\% to 20\%) sample sets.

\section{Results and Analysis}

Computational simulations were performed to assess the accuracy of the proposed explainable tree-based deep learning method (DMR), against other state-of-the-art approaches.

\subsection{Faces Data set}

\vspace{-5pt}

Table \ref{Table1} shows that the proposed DMR method provides the best result in terms of classification accuracy than its state-of-the-art competitors. The number of model parameters for DMR (and xDNN) is, strictly speaking, zero, because the 2 parameters (mean, $\mu$ and standard deviation, $\sigma$) per prototype (\textit{data cloud}) are derived from the data and are not algorithmic parameters or user-defined parameters. However, the tree-based structure of the proposed DMR and the mechanism for balancing the classes allow the result to surpass all others. The propose deep reasoning through a layered pair-wise DT is exploiting and benefiting from the old principle of \textit{divide et impera}. 

\vspace{-5pt}

\begin{table}[H]
	\small \caption{Performance Comparison: Faces-1999 Data set}
	\begin{center}
		\begin{tabular}{c|c}

			\hline
		    \textbf{Method} &\   $Accuracy$\\
			\hline
			DMR & \textbf{\underline{96.71}} \% \\
			VGG--VD--16 & 96.32\%\\
			xDNN   & 96.15\%\\
			VGG--VD--16 & 96.32\%\\
			SVM  & 95.51\% \\
			KNN & 88.54\%  \\
			DT& 61.53\% \\
			\hline
		\end{tabular}
		\label{Table1}
	\end{center}
\end{table}

\subsection{Caltech-101 Data set}

\vspace{-5pt}

Table \ref{Table2} shows the results considering the challenging Caltech-101 data set. It is possible to note through Table \ref{Table2} that the proposed DMR method provides the best result in terms of classification accuracy. The proposed Caltech-101 is hugely unbalanced, and the inner data augmentation mechanism of the proposed DMR method favour the balance of the data, consequently, it improves the final classification result. Moreover, the intelligent tree-based structure of the proposed method allows interpretability and also favours the improvement in the classification accuracy of the given model.  

The proposed explainable tree-based DNN surpasses in terms of accuracy the state-of-the-art VGG--VD--16 algorithm which is a well-established convolutional deep neural network. Moreover, it could also surpass other state-of-art approaches. 

\vspace{-5pt}

\begin{table}[H]
	\small \caption{Performance Comparison: Caltech-101 Data set}
	\begin{center}
		\begin{tabular}{c|c}

			\hline
		    \textbf{Method} & $Accuracy$ \\
			\hline
			DMR & \textbf{\underline{94.31\%}} \\
			SPP-net & 91.44\% \\
			xDNN  & 90.62\%\\
			VGG--VD--16 & 90.32\%\\
			KNN & 85.65\% \\
			DT &54.42\%\\
			\hline
		\end{tabular}
		\label{Table2}
	\end{center}
\end{table}

\subsection{Caltech-256 Data set}

Results for Caltech-256 are presented in Table \ref{Table3}. 

\begin{table}[H]
	\small \caption{Performance Comparison:  Caltech-256 Data set}
	\begin{center}
		\begin{tabular}{c|c}

			\hline
		    Method &$Accuracy$ \\
			\hline
			DMR	& \underline{\textbf{77.54}}\% \\
			xDNN \cite{angelov2019explainable}	& 75.41\% \\
			SVM(1) \cite{zeiler2014visualizing}	& 24.6 \% \\
			SVM(2)	\cite{zeiler2014visualizing}& 39.6\% \\
			SVM(3)	\cite{zeiler2014visualizing}& 46.0\% \\
			SVM(4) \cite{zeiler2014visualizing}& 51.3\% \\
			SVM(5)	\cite{zeiler2014visualizing}& 65.6\%\\
			SVM(7) \cite{zeiler2014visualizing}& 71.7\% \\
			Softmax(5)\cite{zeiler2014visualizing}	& 65.7\% \\
			Softmax(7)	\cite{zeiler2014visualizing}& 74.2\% \\
			\hline
		\end{tabular}
		\label{Table3}
	\end{center}
\end{table}

These results demonstrate that the proposed DMR approach obtains the best classification accuracy ever reported for this complex problem, namely, 77.54\%. The proposed approach not only surpasses all published competitors but also offers a clearly explainable model.

\begin{figure}[H]
	\begin{center}
	IF (Image $\sim$ {\includegraphics[scale=0.28]{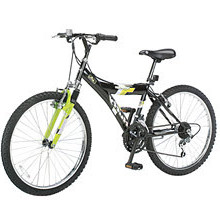}})  OR \nonumber (Image $\sim$ {\includegraphics[scale=0.18]{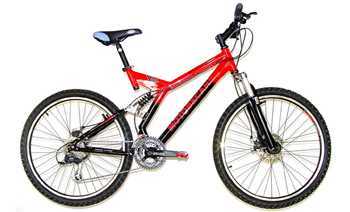}})OR\\ ... OR  (Image $\sim$ {\includegraphics[scale=0.32]{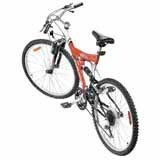}}) \nonumber THEN "Mountain Bike"
		\label{drb_class}
	\end{center}
\end{figure}

 DMR even surpasses the recently introduced by us xDNN approach \cite{angelov2019explainable}, which reported the world best result on 5 December 2019 for this classification problem.

\section{Conclusion}

 In this paper we introduce the DMR -- a prototype-based explainable DNN with DT inference and balanced amount of prototypes per class regardless of the possible imbalances of the training data. The proposed method offers two main novelties, namely: i) using a DT to determine the winning class label, and ii) balancing the classes by synthesising data around the prototypes determined from the available training data. It demonstrates excellent performance surpassing three well known benchmark problems (Caltech-101, Caltech-256 and Faces-1999) where the first two are the the best results published. The proposed approach is explainable-by-design, computationally efficient (no need for GPUs, high degree of parallelization possible, no iterative search procedures and parameter optimisation). Furthermore, it offers the ability to learn continuously (live-long) adapting smoothly to new data patterns. It is a step towards bringing closer machine learning and automated reasoning into what we call \textit{deep machine reasoning} aiming not only high levels of accuracy but also deeper understanding and insight. 
 
 \bibliography{report}   
\bibliographystyle{IEEEtran}

\end{document}